\documentclass[a4paper,fleqn]{cas-sc}

\usepackage[numbers]{natbib}
\usepackage{subfigure}
\usepackage{multirow}
\usepackage{amsmath}
\begin{document}
\shorttitle{Regional Function Guided Traffic Flow Prediction}
\title[mode = title]{Urban Regional Function Guided Traffic Flow Prediction}



\author[1]{Kuo Wang}
\fnmark[1]

\author[2]{LingBo Liu}
\fnmark[1]

\author[1] {Yang Liu}
\cormark[1]

\author[1]{GuanBin Li}

\author[1]{Fan Zhou}

\author[1]{Liang Lin}

\address[1]{School of Computer Science and Engineering, Sun Yat-sen University, China}
\address[2]{Department of Computing, the Hong Kong Polytechnic University, Hong Kong}

\nonumnote{wangk229@mail2.sysu.edu.cn(K. Wang);lingbo.liu@polyu.edu.hk(L. Liu);liuy856@mail.sysu.edu.cn(Y. Liu);liguanbin@mail.sysu.edu.cn(G. Li);isszf@mail.sysu.edu.cn(F. Zhou);linliang@ieee.org(L. Lin)}
\fntext[1]{Equal contribution}
\cortext[1]{Corresponding Author}




\begin{abstract}
The prediction of traffic flow is a challenging yet crucial problem in spatial-temporal analysis, which has recently gained increasing interest. In addition to spatial-temporal correlations, the functionality of urban areas also plays a crucial role in traffic flow prediction. 
\textcolor{black}{However, the exploration of regional functional attributes mainly focuses on adding additional topological structures, ignoring the influence of functional attributes on regional traffic patterns. Different from the existing works, we propose a novel module named POI-MetaBlock, which utilizes the functionality of each region~(represented by Point of Interest distribution) as metadata to further mine different traffic characteristics in areas with different functions.}
Specifically, the proposed POI-MetaBlock employs a self-attention architecture and incorporates POI and time information to generate dynamic attention parameters for each region, which enables the model to fit different traffic patterns of various areas at different times. Furthermore, our lightweight POI-MetaBlock can be easily integrated into conventional traffic flow prediction models. Extensive experiments demonstrate that our module significantly improves the performance of traffic flow prediction and outperforms state-of-the-art methods that use metadata.
\end{abstract}

\maketitle

\section{Introduction}
\textcolor{black}{With the rapid advancement of data analytical technology, the construction of Intelligent Transportation Systems (ITS) has become pervasive \cite{zhang2011data,liu2022causal}. Traffic flow prediction is an essential component of ITS and plays a critical role in urban management and public safety \cite{guo2019attention,zhu2022hybrid}. The primary objective of traffic flow prediction is to forecast the traffic flow in different regions of the city or on roads based on past traffic conditions, such as traffic volume or speed, over a specific period of time.}

\begin{figure}[t]
  \centering
  \subfigure[Diverse correlations of traffic flow. Except for the traditional spatial-temporal correlations, the functionality correlations~(red line) of the region will also affect the traffic flow, even if the regions with similar functions are far away from each other.]{
    \label{fig0:subfiga}
    \includegraphics[width=2.2in]{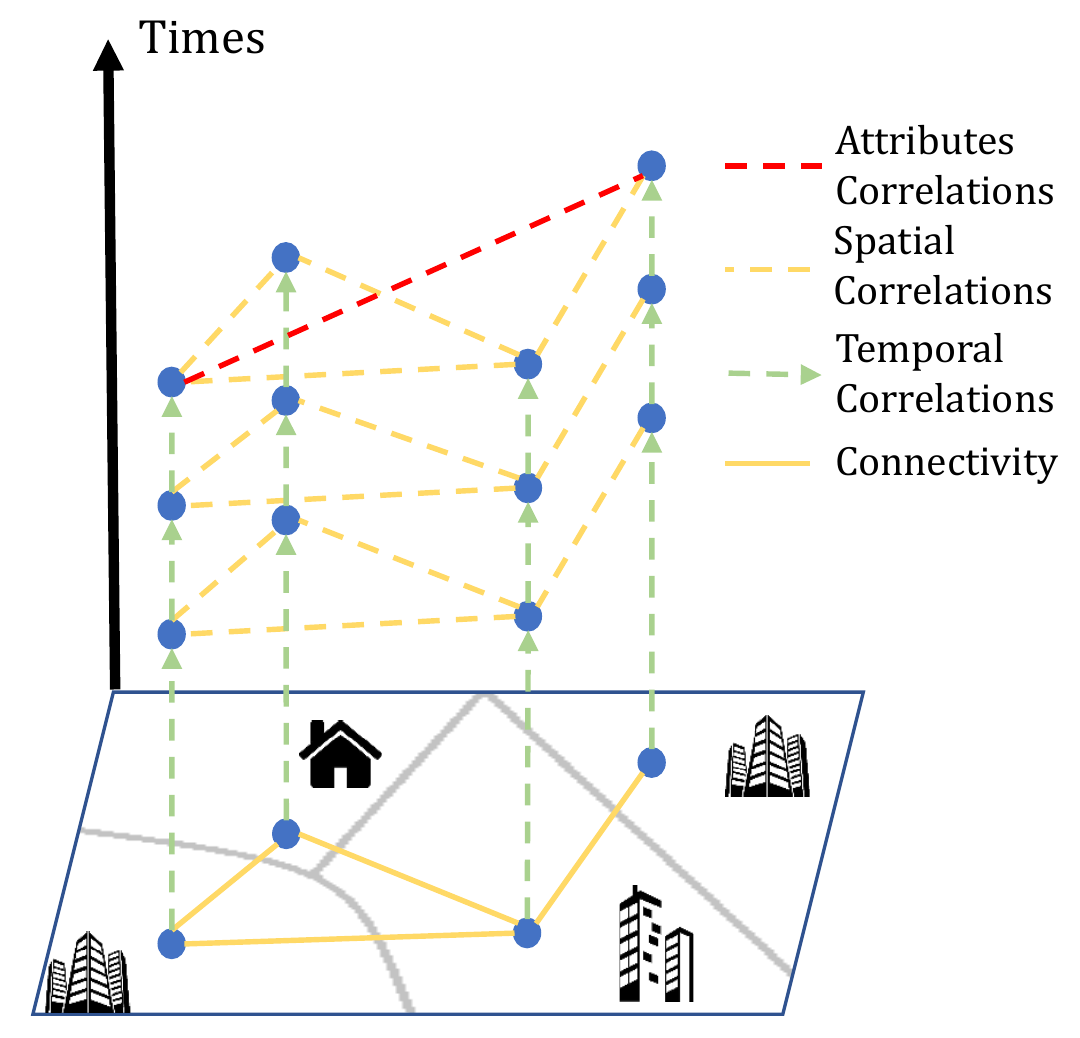}}
  \hspace{0.5in}
  \subfigure[The impact of time on traffic flow. The traffic shows different patterns in different time intervals, moreover, traffic patterns at the same time period on different days are similar. ]{
    \label{fig0:subfigb}
    \includegraphics[width=2.5in]{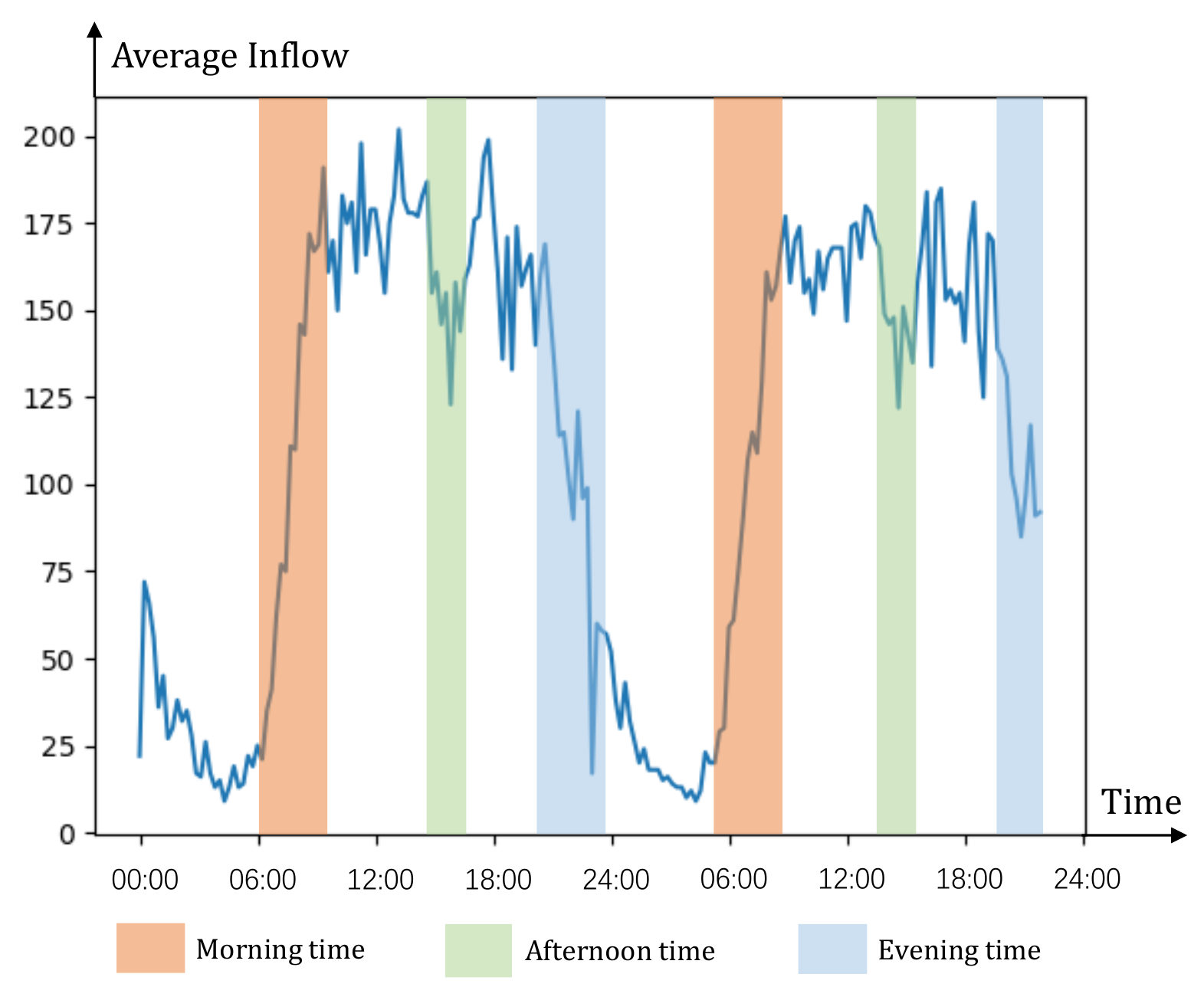}}

  \caption{Complex correlations and influencing factors of traffic flow.}
  \label{fig0}
\end{figure}

\textcolor{black}{Traffic flow prediction is a classic spatial-temporal prediction problem \cite{liu2021temporal,liu2022cross}. Recently, the mainstream approach is combining GCN \cite{defferrard2016convolutional} and RNN \cite{hochreiter1997long} to model spatial-temporal correlation \cite{liu2019deep,liu2021semantics,CHEN2022522,CAO2022185,liu2022tcgl,LIU202381,XU2023580}, such as DCRNN \cite{li2017diffusion}, STGCN \cite{yu2017spatio}. As shown in Figure \ref{fig0}, except for the conventional spatial-temporal correlations \cite{liu2018transferable,liu2018global,liu2018hierarchically}, the traffic flow is also affected by other factors, such as time and regional functionalities. For these additional factors, GMAN \cite{zheng2020gman} proposed to add time stamp information to generate dynamic traffic features. DMVST-Net\cite{yao2018deep}, ST-MGCN\cite{geng2019spatiotemporal} used additional functionality topology to predict taxi and ride-hailing demand. Although the additional factors can improve the accuracy of prediction, the existing traffic flow prediction methods still suffer from the following three challenges:}

\begin{itemize}
\item \textcolor{black}{The traffic patterns in various regions should differ from each other. Nevertheless, existing models primarily employ a single set of parameters to predict traffic flow across different regions within the entire city, disregarding the diverse traffic patterns present in distinct areas.}
\item \textcolor{black}{The influence of functional attributes on traffic changes over time. However, existing models often overlook the dynamic impact of time when exploring the functionality.} 
\item \textcolor{black}{The capacity to integrate regional functionality information is restricted in the proposed network structure. Additionally, there is a lack of a standardized approach for incorporating supplementary meta-information into newly proposed models.}
\end{itemize}

\textcolor{black}{Given the above considerations, we utilize POI (Points of Interest) distribution as a representation of functional attributes of a region and propose a novel module, named POI-MetaBlock, which leverages POI data and time stamps to facilitate traffic flow prediction. The POI-MetaBlock adopts the self-attention structure \cite{vaswani2017attention} and integrates time information to achieve dynamic attention, enabling our module to capture traffic patterns dynamically. We treat the POI distribution as metadata for parameter generation \cite{pan2019urban}, which enables the model to accommodate different traffic patterns across various regions. To model functionality correlations that are overlooked by the original geographic topology, we establish a second topological structure (represented by the red line in Figure \ref{fig0}) using cosine similarity of POI distribution and apply a graph convolution network \cite{kipf2016semi} to refine attention results in each region based on the functionality topological structure. It is noteworthy that the proposed module can be easily integrated into conventional models through a residual connection, allowing existing models to additionally account for the impact of functional attributes on the original basis.}

\textcolor{black}{In commonly used traffic prediction datasets, the urban area is often obtained through the grid method, which can lead to a loss of the original functional areas of the city. To address this limitation, we propose to divide the city into irregular areas based on its road network \cite{sun2020predicting}, and create two traffic flow prediction datasets incorporating POI data. }

In summary, our contributions are mainly summarized into the following aspects:
\begin{itemize}

\item \textcolor{black}{We propose a novel module named POI-MetaBlock, which leverages POI\&time data to dynamically generate prediction parameters and fit traffic patterns of different areas at different times.}
\item The proposed POI-MetaBlock is a lightweight module that can be easily integrated into other traffic flow prediction models. It allows the original model to take advantage of functional correlations during traffic flow prediction.
\item \textcolor{black}{We constructed two new datasets by partitioning the city into irregular areas using POI information. Our experiments demonstrate that our proposed POI-MetaBlock module can achieve a significant improvement in performance over various existing traffic prediction models on these datasets.}
\end{itemize}

\section{Related Works}

\paragraph{Different Types of Traffic Prediction.}
Generally speaking, traffic prediction refers to the prediction of traffic conditions in the future, including flow, speed, vehicle location, and so on. According to the scope, the traffic prediction can be aimed at individuals, roads, and the whole urban areas. For example, \cite{fan2015citymomentum, song2014prediction} predicted an individual's trajectory based on their location history; \cite{abadi2014traffic, xu2014accurate} aimed to predict traffic speed and traffic volume on the road; \cite{chen2014road, hoang2016fccf,9338393} are methods for the city scale regional traffic prediction; In addition, there are some special traffic prediction tasks: \cite{9338315} focused on interpretability, and proposed an interpretable deep learning model for traffic flow prediction. \cite{zhang2019trafficgan} aimed to estimate the traffic after new buildings were constructed. This paper focuses on the urban scale regional traffic prediction and innovatively incorporates regional functionality to guide the traffic flow prediction.

\paragraph{Different Methods for Traffic Prediction.}
In the past, there are many machine learning methods that can be utilized for traffic flow prediction. For example, using support vector regression (SVR) for travel-time prediction \cite{wu2004travel}, using the traditional methods ARIMA \cite{makridakis1997arma} for time series prediction tasks. Later, in view of the excellent fitting ability of RNN~\cite{hochreiter1997long} for sequence-structure data, many RNN-based methods~\cite{sutskever2014sequence} have been proposed to deal with time-series tasks. At the same time, CNN has made great achievements in feature extraction, so combining CNN and RNN for traffic flow prediction \cite{zhang2017deep, zhang2016dnn} becomes popular. \textcolor{black}{Furthermore,  AttConvLSTM~\cite{liu2021modeling} is proposed to combine the attention mechanism for emphasizing the impact of certain parts on traffic flow predictions.} Recently, graph neural network has become a hot topic. \cite{bruna2013spectral} proposed GCN based on the spectral graph theory, then generalized neural networks to graph-structured data becomes common \cite{bronstein2017geometric}. At present, there are two mainstream paradigms to apply GCN to graph-structured data, including spectral-based graph neural network~\cite{li2018adaptive, kipf2016semi}, and spatial GCN-based methods~\cite{atwood2016diffusion}. Since there is a natural graph structure between urban areas, adopting graph neural networks to predict traffic flow has become the mainstream~\cite{yu2017spatio, lv2018lc, li2017diffusion, peng2021dynamic}. \textcolor{black}{STP-TrellisNets~\cite{ou2020stp} further adopts dynamic spatial graphs to predict the metro passenger flows.}
Our POI-MetaBlock adopts graph networks and the self-attention mechanism to assist traffic flow prediction. \textcolor{black}{Different from AttConvLSTM~\cite{liu2021modeling}, our attention module is independent of other network structures while the attention parameters are dynamically generated from the POI distribution of the corresponding areas. Note that the spatial topology and functional relationship in urban traffic flow prediction are fixed, which is different with the condition of metro passenger flow prediction~\cite{ou2020stp}.}

\paragraph{Prediction With External Information.}
There are many existing methods that also utilize external information to assist prediction. External factors here are mainly composed of weather, POI, regional functional attributes, etc. The ways of combining additional information are mainly divided into adding extra topology and expanding the flow features. \cite{geng2019spatiotemporal,yao2018deep,shi2020predicting} add the topology of POI similarity into the original geographic topology to predict the traffic; while \cite{zheng2019deepstd} combines weather conditions and POI data into traffic features to predict traffic flow. The DHSTNet \cite{ali2021exploiting} combines time information to generate dynamic additional embedded information to assist traffic prediction. \textcolor{black}{Similarly, STRN~\cite{liang2021fine} dynamically generates embedded expressions for POI based on external factors such as weather, and combines them with traffic characteristics for prediction.} These models use a single set of parameters to predict traffic in different urban regions which is difficult to explore different traffic patterns between different areas.

Our POI-MetaBlock adopts POI distribution combined with time stamps to model the regional function's dynamic influence on traffic. \textcolor{black}{Different from the above works, instead of generating embedding for external information, we generate independent prediction parameters with metadata of each region, which can better enable the model to capture the different traffic patterns of different regions at different times.} Moreover, our module has strong applicability, which is easy to integrate the module into traditional spatiotemporal prediction networks.

\section{Preliminaries}

\subsection{Irregular Regions}

\textcolor{black}{The city is a complex system that is divided into various irregular parts by the roads, each playing a unique role in the city's functioning, such as residential, leisure, and office areas. Transportation is crucial to people's daily life as they travel between these areas. The regular grid division fails to capture the original functional areas of the city, impeding the discovery of the underlying reasons behind traffic flow changes. To address this issue, we propose dividing the urban area into irregular blocks based on the main roads of the city to forecast traffic flow on these irregular blocks.}

\begin{figure}[t]
  \centering
  \subfigure[Original road map.]{
    \label{fig:subfig:a} 
    \includegraphics[width=1.3in]{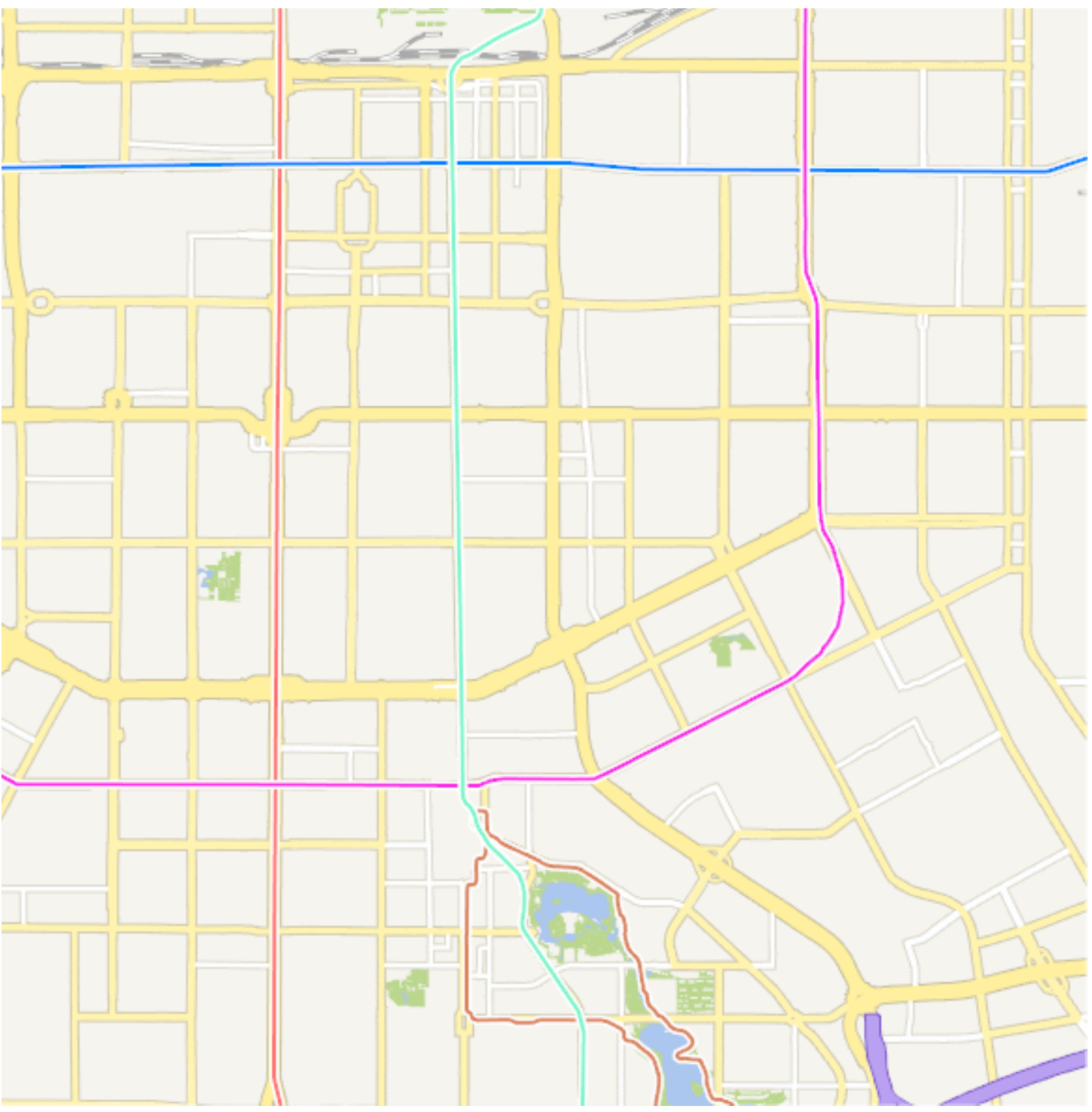}}
  \hspace{0.1in}
  \subfigure[After outline.]{
    \label{fig:subfig:b} 
    \includegraphics[width=1.3in]{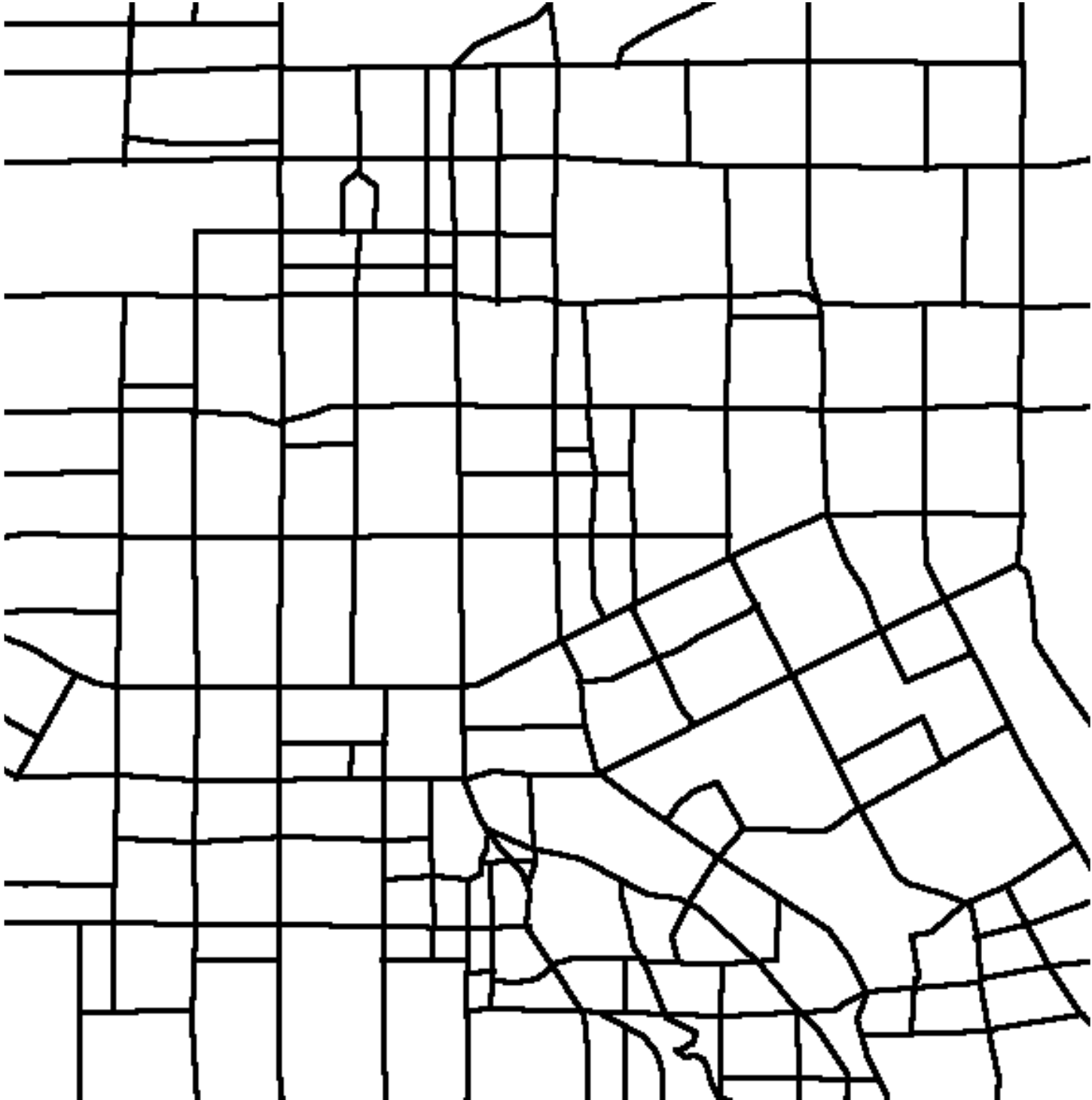}}
  \hspace{0.1in}
  \subfigure[After dilation.]{
    \label{fig:subfig:c} 
    \includegraphics[width=1.3in]{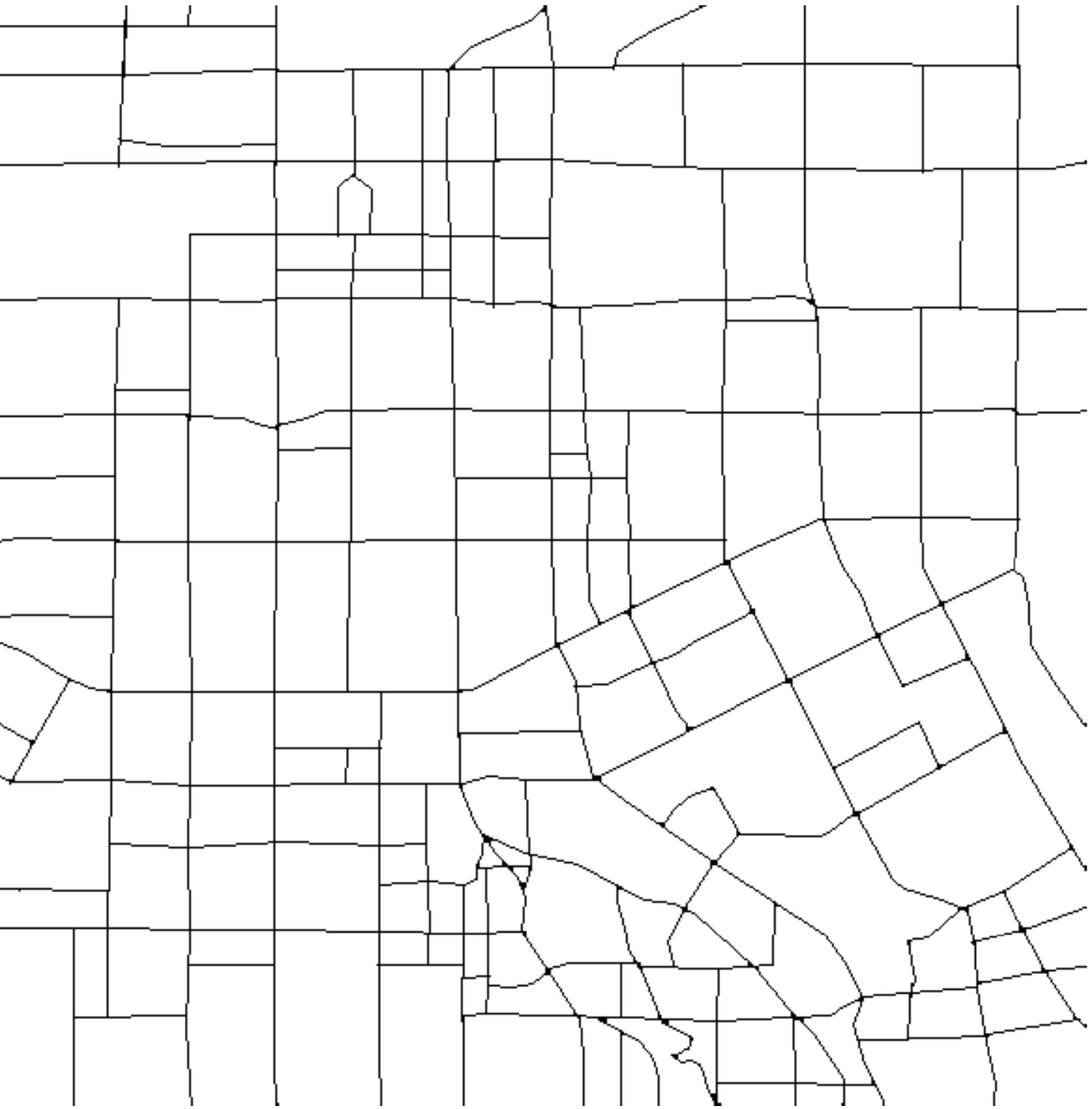}}
  \hspace{0.1in}
  \subfigure[After merging.]{
    \label{fig:subfig:d} 
    \includegraphics[width=1.3in]{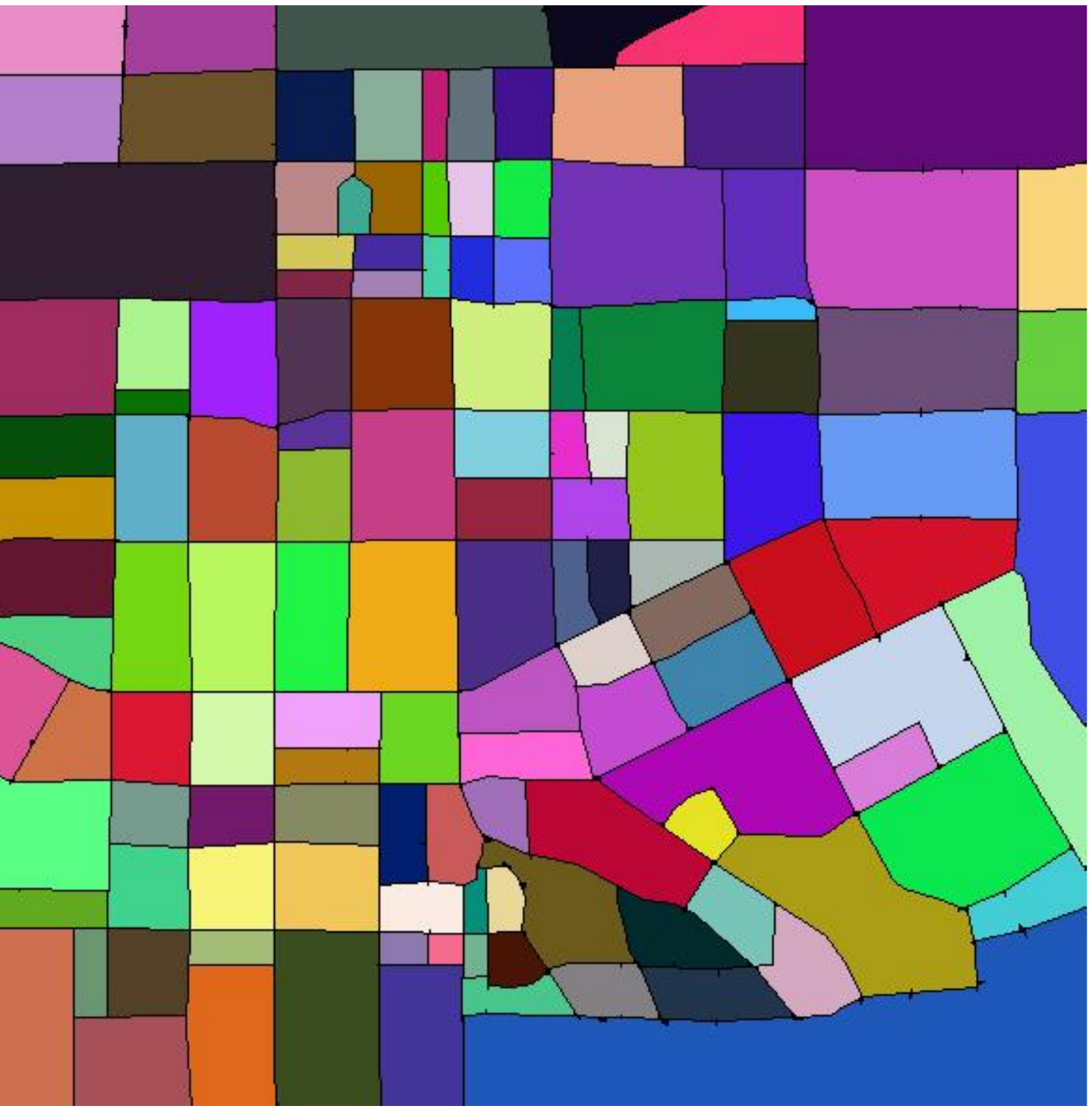}}
  \caption{The process of irregular region division.}
  \label{fig:subfig} 
\end{figure}

\textcolor{black}{Our partitioning process comprises three stages: road extraction, road rendering, and region merging. Initially, we obtain the map of the city's road network, as depicted in Figure \ref{fig:subfig:a}. Subsequently, we highlight the main roads on the map using thick lines. The urban area is represented by the white portion in Figure \ref{fig:subfig:b}, while the black lines represent the road network structure.}

\textcolor{black}{After extracting the road network structure and outlining the main roads, we refine the road using the image dilation algorithm, as illustrated in Figure \ref{fig:subfig:c}. This process yields a preliminary division of the urban area. Next, we identify small and marginal areas with low traffic flow, defined as having less than 10 inflows or outflows in more than 75\% of the time intervals, and merge them with their surrounding regions according to the original map. The final result is obtained by applying the connected component labeling algorithm (CCL), which provides labels for each irregular region, as shown in Figure \ref{fig:subfig:d}. The adjacency attributes between different regions are used as the topological structure for region partition.}

\textcolor{black}{The track data used in this study is sourced from the Didi Chuxing GAIA Initiative\footnote{https://gaia.didichuxing.com}, while the inflow and outflow are calculated based on the track data. Moreover, we collect POI data associated with each region to augment our datasets. The datasets will be publicly available to enable further research into POI information and urban traffic prediction.}

\subsection{Problem Definition}

\textbf{Graph Network}\quad We define the urban area as an undirected graph network $G=(V, E, \mathbf{A})$ after irregular division. A node $v_i\in V$ represents an irregular region. $|V|=N$, which represents there are $N$ regions in the urban area. $E$ is the edge set, which indicates the neighbor relationship between regions. $\mathbf{A} \in \mathbb{R}^{N \times N}$ denotes the adjacency matrix of graph $G$.

\textbf{Traffic Flow}\quad In a time slice, each region of the urban area will generate traffic flow, which can be divided into input flow and output flow, respectively. Since a period of time contains many time slices, the whole urban area's traffic flow information can be represented as $\mathbf{X}\in \mathbb{R}^{N\times T\times D}$, where $N$ is the number of different regions, $T$ is the number of time slices and $D$ is the dimension of traffic information. In our work, a time slice is 15 minutes, and the dimension of traffic information $D$ is set to 1, which means that using inflow to predict inflow, using outflow to predict outflow.

\textbf{Traffic Flow Predicting}\quad Given the traffic flow $\mathbf{X}\in \mathbb{R}^{N\times T\times D}$ in the past period of time, our goal is to predict the traffic flow $\mathbf{Y}\in \mathbb{R}^{N\times T^{\prime}\times D}$ in the next period of time. The two periods of time $T$ and $T^{\prime}$ are continuous and there is no interval or intersection between them. In our work, both $T$ and $T^{\prime}$ are set to 4, which means using the traffic flow information of the past hour to predict the traffic flow information of the next hour.

\section{METHODOLOGIES}

\begin{figure}[t]
  \centering
  \includegraphics[width=4.5in]{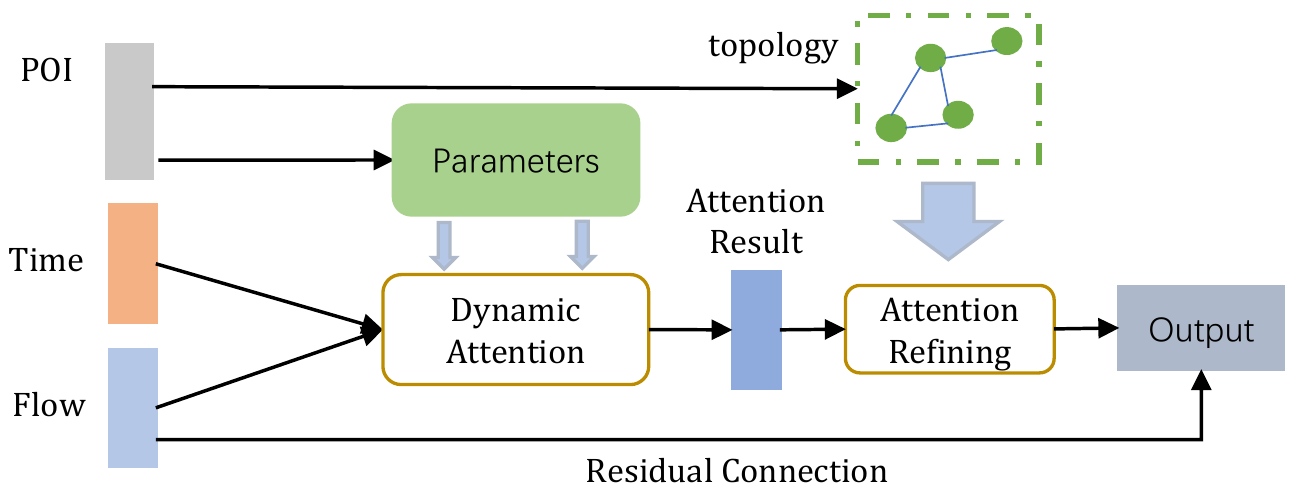}
  \caption{\textcolor{black}{The POI-MetaBlock module consists of three components: dynamic attention, parameter generation, and attention refining. Firstly, the dynamic attention scores are calculated based on the traffic flow features and time stamps. Next, the attention parameters are generated using POI information. Then, the attention result is refined using graph networks based on POI similarity. Finally, a residual connection is applied for the final prediction.}}
  \label{fig1}
\end{figure}

\textcolor{black}{Our approach leverages the distribution of POI data, which can provide insights into the functional attributes of different regions, to enhance traffic flow prediction. To this end, we propose a novel module, named POI-MetaBlock, which integrates traffic flow features, POI distribution, and time stamps to assist the prediction. Our module utilizes Self-Attention architecture \cite{vaswani2017attention} to facilitate traffic flow prediction, as shown in Figure \ref{fig1}. The module consists of three main components: Dynamic Attention, Parameter Generation, and Attention Refining. The Dynamic Attention block calculates dynamic attention scores according to flow features and time stamps, while the parameter is generated from the POI information. The topology corresponding to POI similarity is then used to refine the attention results. Finally, the refined attention results are added to the original traffic features by residual connection \cite{he2016deep} to generate the final prediction, enabling the model to capture the functional correlations of different regions at different times. In the following subsections, we provide a detailed description of the network structure.}

\subsection{Dynamic Attention}
\textcolor{black}{The impact of time on traffic flow is evidently significant. To account for this, we incorporate time information into our module, enabling the dynamic generation of attention scores that capture the ever-evolving nature of traffic patterns. This approach allows us to better understand and model the dynamic characteristics of traffic flows.}

\textcolor{black}{Following \cite{zheng2020gman}, we assign unique identifiers to each time slice in a day, as well as each day of the week. Specifically, our approach regards 15-minute intervals as the unit of time, resulting in a total of 96 distinct daily IDs and 7 weekly IDs. By combining these IDs, we are able to uniquely identify each time slice within a given week. This enables us to more effectively analyze and interpret the data associated with each time period.}

In dynamic attention, the input traffic feature $\mathbf{X}$ is $\mathbb{R}^{N\times T\times d}$, where $d$ represents the dimension of traffic features. Our goal is to predict the traffic feature in the next period of time, which is $\mathbf{X}^{\prime}\in \mathbb{R}^{N\times T^{\prime}\times d}$. Note that $T^{\prime}=T$ in our datasets and experiments. For clarification, we use $T^{\prime}$ to represent the next period of time. In our block, the input of time information is the daily ID and the weekly ID of the time period ($T+T^{\prime}$). First, we use one-hot embedding to transform the ID into a time vector. After that, we concatenate the daily and weekly time vectors together to get the 103-dimensional temporal embedding corresponding to the traffic flow features, which is denoted as $\mathbf{E} \in \mathbb{R}^{(T+T^{\prime})\times 103}$.

\textcolor{black}{To perform the dynamic attention, two full connection layers with the relu activation function is adopted to transform the temporal embedding to size $d$, and then we split it as $\mathbf{E}_1$ and $\mathbf{E}_2$ according to the time sequence, resulting in $\mathbf{E}_1\in \mathbb{R}^{T\times d},\mathbf{E}_2\in \mathbb{R}^{T^{\prime}\times d}$. After that, $\mathbf{E}_1$ and $\mathbf{E}_2$ are concatenated with the traffic features in each region separately to generate traffic features in different time $\mathbf{X}_1\in \mathbb{R}^{N\times T\times 2d}$ and $\mathbf{X}_2\in \mathbb{R}^{N\times T^{\prime}\times 2d}$.} The Dynamic Attention takes traffic features with different time stamps as inputs, and then calculates the dynamic attention score as follows:

For each $X_1\in\mathbf{X}_1$ and $X_2\in\mathbf{X}_2$, $X\in\mathbf{X}$:

\begin{equation}
\begin{aligned}
&Q = X_2W_q,X_2\in\mathbb{R}^{T^{\prime}\times 2d},W_q\in \mathbb{R}^{2d\times d^{\prime}},\\
&K = X_1W_k,X_1\in\mathbb{R}^{T\times 2d},W_k\in \mathbb{R}^{2d\times d^{\prime}},\\
&V = XW_v,X\in\mathbb{R}^{T\times d},W_v\in \mathbb{R}^{d\times d^{\prime}},
\end{aligned}
\end{equation}

\begin{equation}
X^{\prime} = \textrm{softmax}(\frac{QK^T}{\sqrt{(2d)}})V \quad X^{\prime}\in\mathbb{R}^{T^{\prime}\times d^{\prime}}.
\end{equation}

\textcolor{black}{The output of the dynamic attention is denoted as $\mathbf{X^{\prime}}\in \mathbb{R}^{N\times T^{\prime}\times d^{\prime}}$. It is worth noting that the parameters $W_q, W_k,$ and $W_v$ are different for each region and the details will be explained in the following chapter.}

\subsection{Parameter Generation}

\begin{figure}[t]
  \centering
  \includegraphics[width=3.5 in]{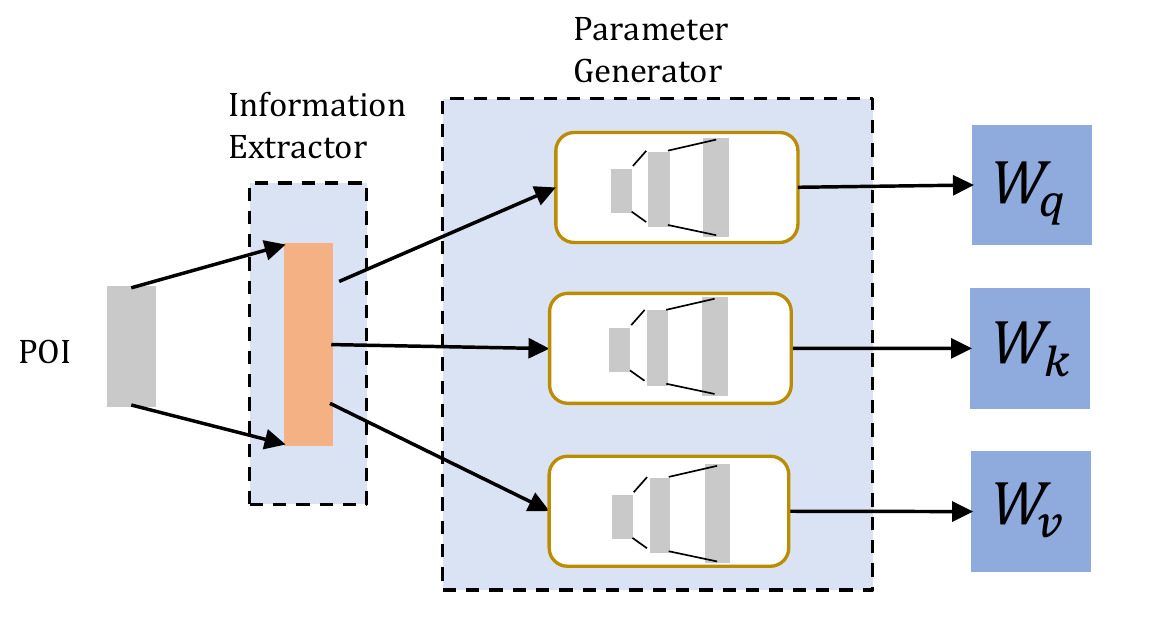}
  \caption{The process of parameter generation.}
  \label{generation}
\end{figure}

\textcolor{black}{Dynamic attention and self-attention models are both limited by the fact that they rely on a single set of attention parameters for all regions. However, as traffic patterns can vary significantly between different functional areas, it is important to account for these differences in our models. To address this issue, we propose to generate unique transformation matrices for each region, based on their specific functionality (as represented by POI data). A detailed illustration of the parameter generation process is provided in Figure \ref{generation}.}

\textcolor{black}{To generate accurate and meaningful transformation matrices for each region, it is crucial to preprocess the raw POI data in a thorough and systematic manner. In our study, we categorize the POI data into 21 distinct categories, which are \{Business Buildings, Bank Finance, Subway Entrance, Educational Schools, Sports and Fitness, Media Organizations, Government Offices, Scenic Spots, Shopping Centers, Hotels, Real Estate Communities, Delicious Food, Residential Districts, Entertainment and Leisure Venues, Bus Stations, Industrial Parks, Medical Care Facilities, Corporate Enterprises, Life Services, Metro Stations\}. These categories were naturally derived from the web server data sources and are representative of the functional attributes of each region. It should be noted that the number of categories can vary depending on the study, as long as they accurately capture the relevant functional characteristics of the regions being analyzed.}

\textcolor{black}{To construct the POI information matrix, we begin by recording the number of POIs of each type in each region. This information is then used to create a POI distribution matrix. To ensure that our results are robust and reliable, we normalize the POI distribution matrix in two ways: (1) line by line, and (2) global mean-std normalization. This normalization process results in two separate matrices, which we then concatenate by columns to obtain the final POI information matrix $P\in \mathbb{R}^{N\times 42}$. }

\textcolor{black}{Our module employs a POI information extractor to identify key elements of POI data, which is subsequently used to generate a transformation matrix for dynamic attention, as shown in Figure \ref{generation}. The POI information extractor comprises a fully connected layer, while the parameter generator utilizes a pyramid multi-layer perceptron (MLP) with two fully connected layers. Specifically, the information extractor transforms the dimension of POI data to 128, with the hidden unit of the first fully connected layer in the MLP set to 256. As conventional conversion matrices can contain both positive and negative values, the activation functions in both the POI information extractor and the parameter generator are set to \textit{tanh}, with no activation function in the final layer of the MLP. To ensure smooth convergence during training, we apply layer normalization to the attention score calculated by $Q$ and $K$. }

\subsection{Attention Refining}
\textcolor{black}{Since the regions with similar functional attributes may not be connected geographically, only using the original topological structure can not capture the functionality correlations between the different regions. To solve this problem, we propose a method to update the attention results by graph convolution on a new topological structure, named attention refining.}

\textcolor{black}{We construct the new topology based on the cosine similarity of POI distribution between regions. The POI information matrix  $P\in \mathbb{R}^{N\times 42}$~(defined in Section 4.2) is adopted for cosine similarity calculation and the final similarity matrix is denoted as $S\in \mathbb{R}^{N\times N}$, where $S_{ij}=\textrm{cos}(P_i, P_j).$} In this new topology, regions with similar POI distribution will be connected. We set a threshold $c$ to get the similarity adjacency matrix $A^{\prime}$. The threshold should not be so large or small, experiments show that thresholds of 0.3--0.6 performs well.
\begin{equation}
{A^{\prime}}_{ij} = \begin{cases}
0& \text{$S_{ij}<c$}\\
1& \text{$S_{ij}\geq c$}
\end{cases}.
\end{equation}

\textcolor{black}{By utilizing the POI similarity adjacency matrix, a novel graph structure can be constructed to enhance the attention results for each region. Let the attention results for the entire urban area be represented by $\mathbf{Z}\in \mathbb{R}^{N\times T^{\prime}\times d^{\prime} }$. The process of refining the graph can be defined as follows.}

\textbf{Graph convolution}\quad In order to make full use of the topological properties of the POI similarity network, at each time slice we adopt graph convolutions\cite{kipf2016semi} based on the spectral graph theory to process the signals.

In spectral graph analysis, the property of the graph is represented by Laplacian matrix $\mathbf{L} = \mathbf{D}-\mathbf{A}$. $\mathbf{A}$ is the adjacency matrix of the graph and $\mathbf{D}$ is the degree matrix. $\mathbf{D}$ is a diagonal matrix, representing the number of nodes connected to each node, ${\mathbf{D}}_{ii}=\sum_{j=1}^N\mathbf{A}_{ij}$. The Laplacian matrix can be symmetric normalized as ${\mathbf{L}}^{sys} =\mathbf{I}_N - \mathbf{D}^{-1/2}\mathbf{L}\mathbf{D}^{-1/2}$.

Fourier transform transforms signals from source domain to spectral domain while inverse Fourier transform transforms signals from spectral domain back to source domain. The convolution process in the source domain can be realized by multiplication in the spectral domain: $f\ast g = \mathcal{F}^{-1}(\mathcal{F}(f)\cdot \mathcal{F}(g))$. On graph structures, the basis of the traditional Fourier transform is a set of eigenvectors of the Laplacian matrix. The eigenvalue decomposition of the Laplacian matrix is $\mathbf{L} = \mathbf{U}\mathbf{\Lambda}\mathbf{U}^T$, where $\mathbf{\Lambda} = diag([\lambda_0,\ldots,\lambda_{N-1}])$ is the eigenvalue matrix and $\mathbf{U} = [u_1,\ldots, u_n]$ is the eigenvector matrix, which is the Fourier basis. In attention refining, the signal of the graph network is the attention results $\mathbf{Z}\in \mathbb{R}^{N\times T^{\prime}\times d^{\prime} }$. Let $z\in \mathbb{R}^{d^{\prime}}$ as the signal of a node at a certain time slice, the graph Fourier transform of $z$ is $\hat{z} = \mathbf{U}^Tz$. $\mathbf{U}$ is the eigenvector matrix of the Laplacian matrix, which means that $\mathbf{U}$ is an orthogonal matrix, so the inverse Fourier transform of $\hat{z}$ is $z = \mathbf{U}\hat{z}$. With a filter $g_{\theta} = diag(\theta)$ parameterized by $\theta \in \mathbb{R}^N$ in the Fourier domain, the convolution on the graph can be defined as:
\begin{equation}
g_{\theta}\ast z = g_{\theta}(\mathbf{L})z = g_{\theta}(\mathbf{U}\Lambda \mathbf{U}^T)z = \mathbf{U}g_{\theta}(\mathbf{\Lambda})\mathbf{U}^Tz.
\end{equation}

Since the eigenvalue decomposition of the Laplacian matrix is very time-consuming when graph is large, in this paper, we use Chebyshev polynomials to approximate the convolution process\cite{simonovsky2017dynamic}, which is:
\begin{equation}
g_\theta\ast z = g_\theta(\mathbf{L})z \approx \sum_{k=0}^{K-1}\theta_kT_k(\mathbf{\tilde{L}})z,
\end{equation}
$\theta_k$ is the $k$th value of parameter $\theta \in \mathbb{R}^K$, and $\mathbf{\tilde{L}} = \frac{2}{\lambda_{max}}\mathbf{L}-\mathbf{I}_N$, $\lambda_{max}$ is the maximum eigenvalue of the Laplacian matrix. $T_k$ is the Chebyshev polynomials, $T_k(z) = 2zT_{k-1}(z) - T_{k-2}(z)$, and $T_0(z)=1, T_1(z) = z$. Using K-order Chebyshev polynomials to approximate the convolution process means using the filter $g_\theta$ to extract the information of 0 to K-1 order neighbors of each node. Finally, the ReLU activation function is used to achieve nonlinearity.

\textbf{Graph Attention}\quad To further enable the model to dynamically update the dependency relationship between neighbors according to the functionality of each region, we use POI information to calculate an additional attention score for the graph convolution process. Suppose the POI information is $P\in \mathbb{R}^{N\times p}$, $p$ is the dimension of the POI data in each region~($p=42$ in our work, as defined in Section 4.2). First, using a fully connected layer with relu activate function to extract the important information $P^\prime \in \mathbb{R}^{N\times d}$.
And then adopting the self-attention mechanism to calculate the attention score:
\begin{equation}
\begin{aligned}
&Q = P^\prime W_q\quad K = P^\prime W_k\quad W_q,W_k\in \mathbb{R}^{d\times d},\\
&S = \textrm{softmax}(QK^T)\quad S\in \mathbb{R}^{N\times N}.
\end{aligned}
\end{equation}

We add the attention score $S$ into the convolution process by accompanying each Chebyshev polynomial $T_k(\mathbf{\tilde{L}})$ with $S$. Using Hadamard product, the convolution process changes to:
\begin{equation}
g_\theta \ast z = \sum_{k=0}^{K-1}\theta_k(T_k(\mathbf{\tilde{L}})\odot S)z.
\end{equation}

\textcolor{black}{To improve the accuracy of traffic flow prediction, we employ attentive graph convolution to refine the attention results of dynamic attention for each region. This method allows regions with similar functional attributes to obtain comparable attention results, thereby enhancing prediction accuracy.}

\subsection{Residual Connection}

\begin{figure}[t]
  \centering
  \includegraphics[width=5.0in]{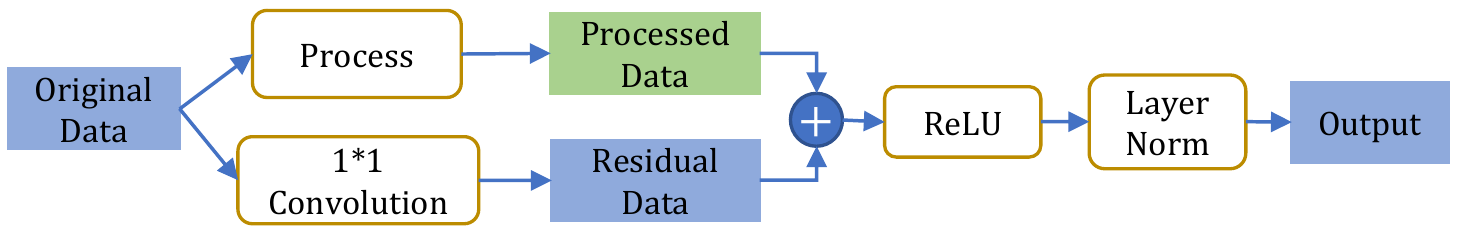}
  \caption{The structure of the Residual Connection. Different from traditional ``$y = x + f(x)$", we use a $1\times 1$ convolution layer to extract information from original data, and add relu activation and layer normalization to the addition result.}
  \label{res}
\end{figure}

\textcolor{black}{In our study, we incorporate residual connections \cite{he2016deep} to enhance attention refinement and optimize the overall structure. The residual connection effectively addresses the issue of vanishing gradient, thereby improving model convergence. In attention refining, the residual connection is computed similarly to that in the POI-MetaBlock. Initially, a $1\times 1$ filter convolutes the input data, and the output is added to the processed data. Following the relu activation function, the resulting values undergo layer normalization to obtain the final output. Figure \ref{res} illustrates the structure.}


\textcolor{black}{The residual connection is derived in two aspects: refine the attention results in the POI-MetaBlock and incorporate the POI-MetaBlock to the conventional networks. For attention refining, the original data in Figure~\ref{res} refers to the attention results in each region while the processed data means the graph conventional results based on the topology of functional similarity. For network incorporation, the original data means the traffic features and the ``process" is our POI-MetaBlock. With the help of residual connection, the model can gain the ability of function perception on the original network, which leads to a more accurate prediction of traffic flow.}

\subsection{Integration with other models}

\begin{figure}[h]
  \centering
  \includegraphics[width=4.5in]{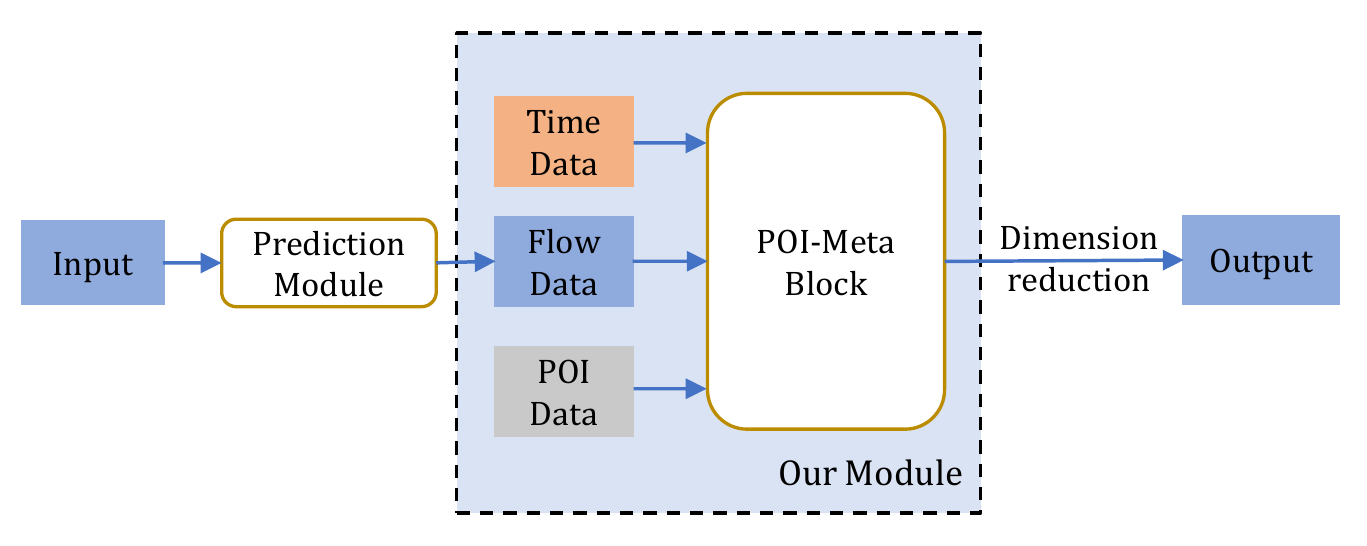}
  \caption{The POI-MetaBlock can be integrated with other models easily. The ``prediction Module" indicates the conventional traffic flow prediction networks.}
  \label{six}
\end{figure}

\textcolor{black}{Our proposed module exhibits excellent flexibility and can be effortlessly integrated into other traffic flow prediction models, as illustrated in Figure \ref{six}. In typical traffic flow prediction models, a fully connected layer or convolution layer is utilized to obtain the final output. By inputting traffic data into our model before dimension reduction and subsequently providing additional time and POI information, we can obtain traffic prediction results that are guided by the functional attributes of urban regions. Our experiments confirm that the inclusion of our POI-MetaBlock module leads to a substantial improvement in model performance.}

\section{Experiments}

\subsection{Datasets}

\textbf{Xi'an Dataset}\quad \textcolor{black}{The Xi'an Dataset comprises 130 distinct regions, with a time period spanning from October 1, 2016 to November 31, 2016. We use 15 minutes as the time unit to calculate traffic information and divide it into inflow and outflow. On average, the dataset contains 75 instances of inflow and 73 instances of outflow per region per time unit.}

\textbf{Chengdu Dataset}\quad \textcolor{black}{The Chengdu Dataset consists of 118 distinct regions, covering the time period from October 1, 2016 to November 31, 2016. We use 15 minutes as the time unit to tally traffic information, dividing it into inflow and outflow. On average, the dataset contains 47 instances of inflow and 46 instances of outflow per region per time unit.}

\begin{figure}[h]
  \centering
  \subfigure[Map-Xi'an]{
    \includegraphics[width=1.0in]{./pic/0.pdf}}
  \hspace{0.2in}
  \subfigure[Regions-Xi'an]{
    \includegraphics[width=1.0in]{./pic/3.pdf}}
\hspace{0.2in}
  \subfigure[Map-Chengdu]{
    \includegraphics[width=1.0in]{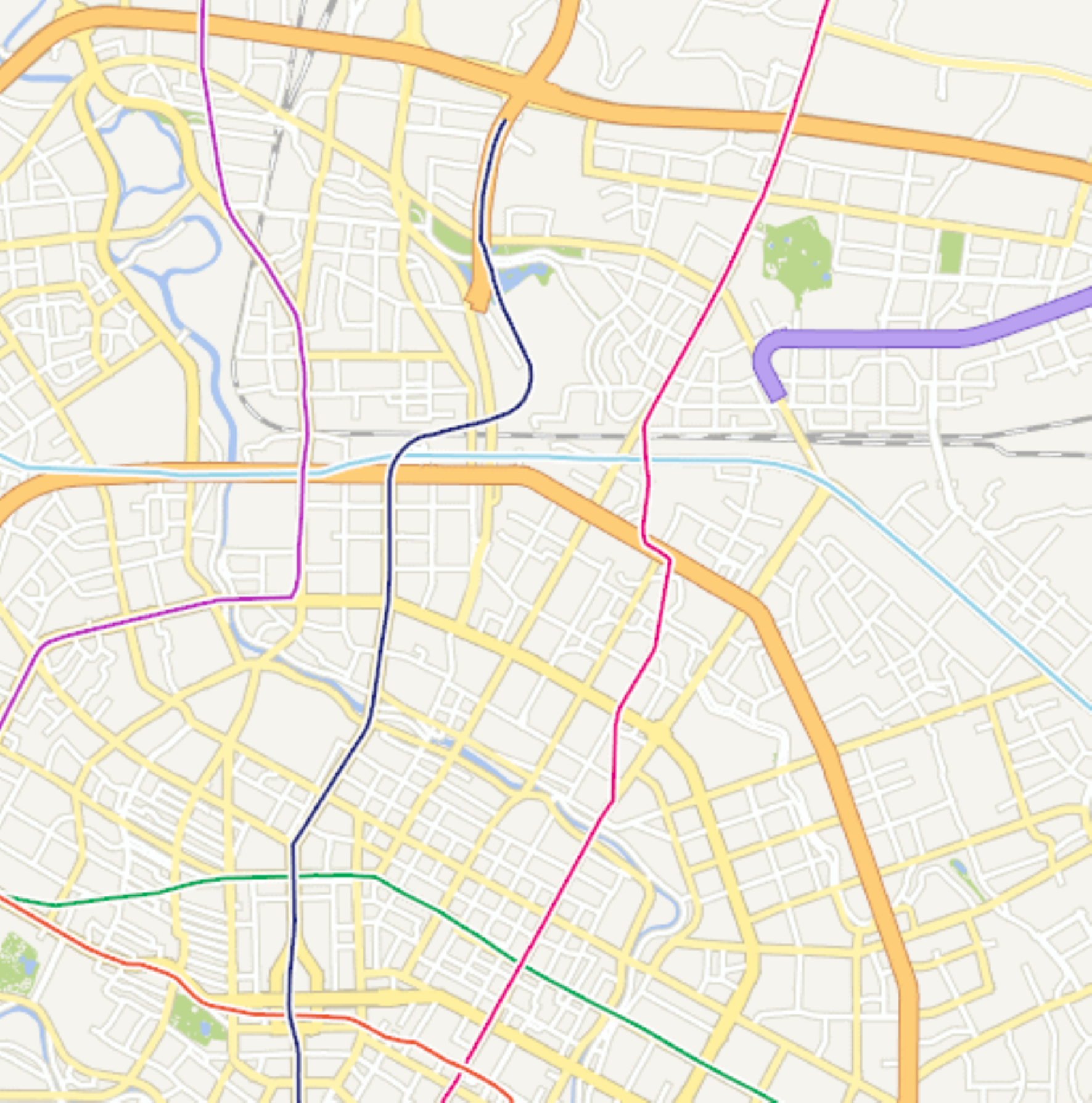}}
  \hspace{0.2in}
  \subfigure[Regions-Chengdu]{
    \includegraphics[width=1.0in]{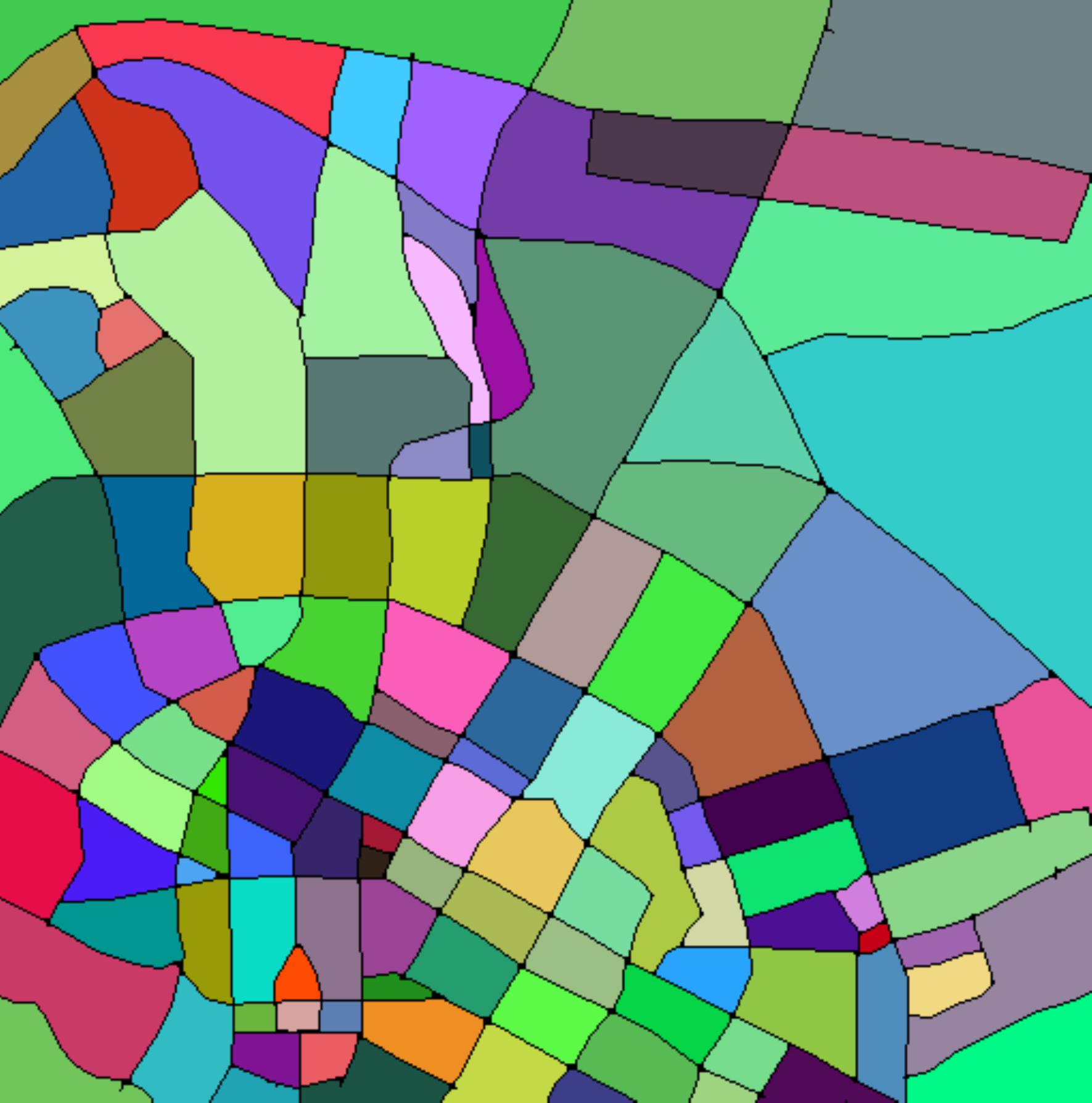}}
  \caption{The map and the divided regions of Xi'an Dataset and Chengdu Dataset.}
  \label{xian} 
\end{figure}

\textcolor{black}{For the two datasets mentioned above, we normalize the input data using mean-std normalization, while keeping the target data in its original form. Using traffic information from the last hour, we predict the inflow and outflow of each region in the next hour. We only use inflow information to predict future inflow and likewise for outflow. Thus, if the input is inflow information, the model outputs predictions for inflow in the next hour, and if the input is outflow information, the model predicts outflow.}

\subsection{Experimental Settings}
\textbf{Baselines.} \quad \textcolor{black}{Our proposed module is integrated into the following 5 baselines: \textbf{DCRNN} \cite{li2017diffusion}, \textbf{STGCN} \cite{yu2017spatio}, \textbf{ASTGCN} \cite{guo2019attention}, \textbf{GAT-Seq2Seq} \cite{pan2019urban}, and \textbf{GMAN} \cite{zheng2020gman}. Among these, \textbf{GMAN} is currently the state-of-the-art model for traffic flow prediction. Compared to other baselines, \textbf{GMAN} additionally considers the time factor and the graph node embedding, enabling the model to overcome geographical restrictions. We compare our module with \textbf{ST-MetaNet} \cite{pan2019urban}, which is the current state-of-the-art model that utilizes meta-information for prediction parameter generation. The structure of \textbf{ST-MetaNet} is based on \textbf{GAT-Seq2Seq}.}

\subsection{Implementation Details.}
\textcolor{black}{To integrate our proposed module into the baseline models, we made the following modifications:}

\begin{itemize}
\item \textbf{DCRNN\footnote{https://github.com/xlwang233/pytorch-DCRNN}:} \textcolor{black}{We set the hidden size to 32, the number of RNN layers to 2, and the diffusion step to 2. To integrate our module, we removed the final full connection layer in the decoder, which is used for dimension reduction, and sent the output of DCRNN, POI data, and time data into our module to obtain the final output.}

\item \textbf{STGCN\footnote{https://github.com/FelixOpolka/STGCN-PyTorch}:} \textcolor{black}{We set the hidden size and output channels to 32 and integrated our module by sending the output of STGCN, POI data, and time data into our module to obtain the final output.}

\item \textbf{ASTGCN\footnote{https://github.com/guoshnBJTU/ASTGCN-r-pytorch}:} \textcolor{black}{We set the hidden size to 32 and the number of blocks to 2. We removed the final convolution layer in ASTGCN, which is typically used for dimension reduction, and sent the output of ASTGCN, POI data, and time data into our module to obtain the final output.}


\item \textbf{GAT-Seq2Seq:} \textcolor{black}{We extended the Seq2Seq model by incorporating a graph attention network layer into both the encoder and decoder. The graph attention mechanism used in this layer is the same as the one used in ASTGCN and is added between the RNN layers. The hidden size of this model remains 32. We then send the output of GAT-Seq2Seq, along with the POI and time data, into our module to obtain the final output.}

\item \textbf{ST-MetaNet\footnote{https://github.com/panzheyi/ST-MetaNet}:} \textcolor{black}{The code was rewritten using PyTorch and the hidden size was set to 32. To replace the meta-information used by ST-MetaNet, we incorporated POI information. We also removed the length information of edges due to the difficulty in measuring the exact distance between irregular regions.}

\item \textbf{GMAN\footnote{https://github.com/zhengchuanpan/GMAN}:} \textcolor{black}{We set the number of attention heads to 4 and the number of STAtt blocks to 1. We removed the final fully connected layer and fed the output of GMAN, along with POI data and time data, into our module to obtain the final output.}

\end{itemize}

\textcolor{black}{In our block, we set the feature dimension of the output of dynamic attention to 16 and use a 1$\times$1 convolution layer to reduce the dimension to 1. We set the output dimension of the POI information extractor to 128 and the dimension of the first full connection layer in the parameter generator to 256. The output dimension of our module is set to 16. We use Chebyshev polynomials of order 3. For all experiments, we use Adam as the optimizer, set the batch size to 64, and set the learning rate to 0.01 initially. After training for 75 epochs, we reduce the learning rate to 0.001. We train the model for a total of 150 epochs using an NVIDIA GeForce GTX 1060 6GB graphics board. The loss function used in all experiments is the MSE loss.}
\begin{equation}
Loss(\mathbf{x},\mathbf{y}) = \frac{1}{n} \sum_{i=1}^{n}(x_{i}-y_{i})^{2}.
\end{equation}

\textbf{Metrics.}\quad Mean absolute error (MAE), rooted mean square error (RMSE) and mean absolute percentage error(MAPE) are adopted to evaluate the performance of our POI-MetaBlock,
\begin{equation}
\mathrm{MAE}=\frac{1}{n} \sum_{i=1}^{n}\left|y_{i}-\hat{y}_{i}\right|, \quad
\mathrm{RMSE}=\sqrt{\frac{1}{n} \sum_{i=1}^{n}\left(y_{i}-\hat{y}_{i}\right)^{2}}, \quad
\mathrm{MAPE}=\sum_{i=1}^{n}\left|\frac{y_{i}-\hat{y}_{i}}{y_{i}}\right| \times \frac{100}{n}.
\end{equation}
where $n$ is the number of regions, $\hat{y}_i$ is the prediction result and $y_i$ is the ground truth.

\subsection{Performance Results}

\begin{table}[!htbp]
\begin{center}
\newcommand{\tabincell}[2]{\begin{tabular}{@{}#1@{}}#2\end{tabular}}  
  \caption{Performance Results on Chengdu Dataset and Xi'an Dataset.}
  \label{tab:result1}
\resizebox{16cm}{!} {
  \begin{tabular}{c|c|ccc|ccc|ccc|ccc}
  \toprule
    \multicolumn{2}{c}{ \multirow{3}*{Models} }& \multicolumn{6}{|c|}{Chengdu}& \multicolumn{6}{|c}{Xi'an}\\
    \cline{3-14}
	\multicolumn{2}{c|}{}&\multicolumn{3}{c|}{Original}&\multicolumn{3}{c|}{POI-MetaBlock}&\multicolumn{3}{c|}{Original}&\multicolumn{3}{c}{POI-MetaBlock}\\
	\cline{3-14}
	\multicolumn{2}{c|}{}&MAE&RMSE&MAPE(\%)&MAE&RMSE&MAPE(\%)&MAE&RMSE&MAPE(\%)&MAE&RMSE&MAPE(\%)\\
	\hline
	\multirow{4}*{DCRNN}&15 min&9.11&13.44&25.07&\textbf{8.46}&\textbf{12.28}&\textbf{20.64}&6.36&9.22&27.32&\textbf{6.05}&\textbf{8.73}&\textbf{23.98}\\
	\multicolumn{1}{c|}{}&30 min&9.61&14.24&25.40&\textbf{8.89}&\textbf{13.02}&\textbf{20.96}&6.85&10.05&28.26&\textbf{6.44}&\textbf{9.41}&\textbf{24.52}\\
	\multicolumn{1}{c|}{}&60 min&10.98&16.39&27.40&\textbf{9.84}&\textbf{14.61}&\textbf{22.01}&7.90&11.77&31.44&\textbf{7.19}&\textbf{10.67}&\textbf{26.14}\\
	\multicolumn{1}{c|}{}&overall&9.96&14.83&25.97&\textbf{9.13}&\textbf{13.44}&\textbf{21.19}&7.12&10.54&29.11&\textbf{6.62}&\textbf{9.74}&\textbf{24.86}\\
	\hline
	\multirow{4}*{STGCN}&15 min&10.21&15.03&27.09&\textbf{9.15}&\textbf{13.18}&\textbf{21.90}&6.63&9.65&28.61&\textbf{6.35}&\textbf{9.13}&\textbf{26.12}\\
	\multicolumn{1}{c|}{}&30 min&11.10&16.45&28.41&\textbf{9.66}&\textbf{14.02}&\textbf{22.15}&7.37&10.89&30.63&\textbf{6.74}&\textbf{9.79}&\textbf{26.63}\\
	\multicolumn{1}{c|}{}&60 min&13.32&20.10&34.00&\textbf{10.70}&\textbf{15.65}&\textbf{23.94}&9.00&13.47&37.78&\textbf{7.67}&\textbf{11.32}&\textbf{28.63}\\
	\multicolumn{1}{c|}{}&overall&11.36&17.54&30.06&\textbf{9.92}&\textbf{14.47}&\textbf{22.70}&7.81&11.63&32.68&\textbf{6.99}&\textbf{10.23}&\textbf{27.19}\\
	\hline
	\multirow{4}*{ASTGCN}&15 min&9.41&13.81&25.37&\textbf{8.54}&\textbf{12.37}&\textbf{23.13}&6.75&9.78&29.39&\textbf{6.26}&\textbf{9.09}&\textbf{24.23}\\
	\multicolumn{1}{c|}{}&30 min&10.46&15.41&28.92&\textbf{8.95}&\textbf{13.11}&\textbf{23.58}&7.59&10.99&33.41&\textbf{6.68}&\textbf{9.83}&\textbf{24.99}\\
	\multicolumn{1}{c|}{}&60 min&12.81&19.27&38.01&\textbf{10.09}&\textbf{15.06}&\textbf{25.39}&9.22&13.29&44.20&\textbf{7.51}&\textbf{11.24}&\textbf{27.50}\\
	\multicolumn{1}{c|}{}&overall&11.06&16.54&31.38&\textbf{9.26}&\textbf{13.67}&\textbf{24.12}&7.99&11.63&36.32&\textbf{6.88}&\textbf{10.20}&\textbf{25.65}\\
	\hline
	\multirow{4}*{GAT-Seq2Seq}&15 min&9.15&13.81&23.03&\textbf{8.44}&\textbf{12.18}&\textbf{20.65}&6.47&9.42&26.68&\textbf{6.23}&\textbf{9.03}&\textbf{24.88}\\
	\multicolumn{1}{c|}{}&30 min&9.87&15.09&24.47&\textbf{8.93}&\textbf{13.00}&\textbf{21.53}&6.94&10.24&28.25&\textbf{6.61}&\textbf{9.72}&\textbf{25.21}\\
	\multicolumn{1}{c|}{}&60 min&11.57&18.25&27.54&\textbf{9.94}&\textbf{14.67}&\textbf{22.98}&7.90&11.76&29.82&\textbf{7.35}&\textbf{10.97}&\textbf{27.02}\\
	\multicolumn{1}{c|}{}&overall&10.30&16.01&24.99&\textbf{9.18}&\textbf{13.44}&\textbf{21.83}&7.17&10.62&28.32&\textbf{6.78}&\textbf{10.03}&\textbf{25.63}\\
	\hline
	\multirow{4}*{GMAN}&15 min&\textbf{8.35}&\textbf{12.01}&20.85&8.47&12.27&\textbf{20.75}&6.06&\textbf{8.68}&24.43&\textbf{6.04}&8.70&\textbf{23.62}\\
	\multicolumn{1}{c|}{}&30 min&8.81&12.88&21.34&\textbf{8.80}&\textbf{12.85}&\textbf{20.86}&6.41&9.33&24.89&\textbf{6.35}&\textbf{9.25}&\textbf{23.73}\\
	\multicolumn{1}{c|}{}&60 min&9.72&14.48&22.48&\textbf{9.60}&\textbf{14.13}&\textbf{21.49}&7.12&10.53&26.77&\textbf{7.05}&\textbf{10.39}&\textbf{25.59}\\
	\multicolumn{1}{c|}{}&overall&9.05&13.31&21.65&\textbf{9.01}&\textbf{13.19}&\textbf{21.04}&6.58&9.64&25.39&\textbf{6.53}&\textbf{9.56}&\textbf{24.34}\\
	
  \bottomrule
\end{tabular}
}
\end{center}
\end{table}

\begin{table}[t]
\begin{center}
\newcommand{\tabincell}[2]{\begin{tabular}{@{}#1@{}}#2\end{tabular}}  
  \caption{Compared with State-of-the-art method: ST-MetaNet.}
  \label{tab:result2}
  \begin{tabular}{c|c|cccc|cccc}
   \toprule
    \multicolumn{2}{c}{ \multirow{2}*{Models} }& \multicolumn{4}{|c|}{Chengdu}& \multicolumn{4}{|c}{Xi'an}\\
    \cline{3-10}
	\multicolumn{2}{c|}{}&15 min&30 min&60 min&overall&15 min&30 min&60 min&overall\\
	\hline
	\multirow{2}*{GAT-Seq2Seq}&MAE&9.15&9.87&11.57&10.30&6.47&6.94&7.90&7.17\\
	\multicolumn{1}{c|}{}&RMSE&13.81&15.09&18.25&16.01&9.42&10.24&11.76&10.62\\
	\hline
	\multirow{2}*{ST-MetaNet}&MAE&9.02&9.58&10.82&9.89&6.45&\textbf{6.89}&\textbf{7.79}&\textbf{7.12}\\
	\multicolumn{1}{c|}{}&RMSE&13.29&14.19&16.15&14.71&9.42&10.18&\textbf{11.67}&10.58\\
	\hline
	\multirow{2}*{\tabincell{c}{POI-Meta\\(No Time)}}&MAE&\textbf{8.84}&\textbf{9.37}&\textbf{10.31}&\textbf{9.59}&\textbf{6.42}&6.92&7.84&7.13\\
	\multicolumn{1}{c|}{}&RMSE&\textbf{13.09}&\textbf{13.90}&\textbf{15.55}&\textbf{14.36}&\textbf{9.33}&\textbf{10.17}&11.70&\textbf{10.56}\\
	\hline
	\multirow{2}*{POI-Meta}&MAE&\textbf{\underline{8.44}}&\textbf{\underline{8.93}}&\textbf{\underline{9.94}}&\textbf{\underline{9.18}}&\textbf{\underline{6.23}}&\textbf{\underline{6.61}}&\textbf{\underline{7.35}}&\textbf{\underline{6.78}}\\
	\multicolumn{1}{c|}{}&RMSE&\textbf{\underline{12.18}}&\textbf{\underline{13.00}}&\textbf{\underline{14.67}}&\textbf{\underline{13.44}}&\textbf{\underline{9.03}}&\textbf{\underline{9.72}}&\textbf{\underline{10.97}}&\textbf{\underline{10.03}}\\

  \bottomrule
\end{tabular}
\end{center}
\end{table}

The performance of the baseline models with or without adding our module on the two datasets is shown in Table \ref{tab:result1}. We can see that after adding our module, the performance of the baseline models is improved significantly for both long-term and short-term forecasts. Except for GMAN, our module bring them 8.3\%--15.1\% performance improvement on Chengdu dataset and 5.4\%--13.1\% performance improvement on Xi'an dataset. For GMAN, our module still makes improvements by 1\%, especially for long term forecasts.

The reason why our module can bring significant performance improvement can be summarized into three aspects:
\begin{enumerate}
\item \textcolor{black}{The traditional traffic flow prediction models usually neglect the impact of time stamps on traffic flow, whereas our proposed module takes the time factor into account and performs dynamic attention based on the time stamps.}

\item \textcolor{black}{In contrast to the traditional model, where all regions share the same set of parameters, our module is capable of generating parameters that are specific to each region. This enables our module to better capture the unique traffic characteristics of each region, resulting in improved forecasting performance.}

\item \textcolor{black}{The traditional models often overlook the fact that similar functional areas have similar traffic characteristics. However, our proposed module builds a new topology based on the similarity of POI information, which enables the model to capture attribute correlations between different regions, even if they are far away from each other. }
\end{enumerate}

\textcolor{black}{Table \ref{tab:result2} illustrates that our module can achieve higher performance gains by better utilizing metadata (POI information), especially on the Chengdu dataset, even without time information. This improvement can be attributed to two main factors. Firstly, ST-MetaNet reduces the dimension of meta information to 2, which limits the module's ability to explore the metadata, while our module preserves the original dimension. Secondly, our module employs self-attention to model time series, while ST-MetaNet still uses GRU, which is another reason why our model outperforms ST-MetaNet. Moreover, self-attention enables easy integration of time information to achieve dynamic attention mechanism.}

\subsection{Ablation Experiments}

\begin{table}[h]\renewcommand\tabcolsep{8pt}
\begin{center}
\newcommand{\tabincell}[2]{\begin{tabular}{@{}#1@{}}#2\end{tabular}}  
  \caption{Performance Results of Ablation Experiments.}
  \label{tab:result3}
  \begin{tabular}{c|ccc|cc|cc|cc|cc}
  \toprule
    \multicolumn{1}{c|}{\multirow{2}*{Datasets}}&\multicolumn{1}{c}{\multirow{2}*{DA}}&\multicolumn{1}{c}{\multirow{2}*{PG}}&\multicolumn{1}{c}{\multirow{2}*{AR}}&\multicolumn{2}{|c}{DCRNN}&\multicolumn{2}{|c}{STGCN}&\multicolumn{2}{|c}{ASTGCN}&\multicolumn{2}{|c}{GAT-Seq2Seq}\\
    \cline{5-12}
	\multicolumn{1}{c|}{}&\multicolumn{3}{c|}{}&MAE&RMSE&MAE&RMSE&MAE&RMSE&MAE&RMSE\\
	\hline
	\multirow{4}*{Chengdu}&-&-&-&9.96&14.83&11.69&17.54&11.06&16.54&13.42&20.52\\
	\multicolumn{1}{c|}{}&\checkmark&-&-&9.48&14.00&11.42&16.81&11.61&17.00&9.44&13.94\\
	\multicolumn{1}{c|}{}&\checkmark&\checkmark&-&9.37&13.91&11.27&16.74&10.18&15.14&9.28&13.78\\
	\multicolumn{1}{c|}{}&\checkmark&\checkmark&\checkmark&\textbf{9.13}&\textbf{13.44}&\textbf{9.92}&\textbf{14.47}&\textbf{9.26}&\textbf{13.67}&\textbf{9.18}&\textbf{13.44}\\

	\hline
	\multirow{4}*{Xi'an}&-&-&-&7.12&10.54&7.81&11.63&7.99&11.63&9.28&13.79\\
	\multicolumn{1}{c|}{}&\checkmark&-&-&6.89&10.18&7.64&11.20&8.07&11.70&7.07&10.30\\
	\multicolumn{1}{c|}{}&\checkmark&\checkmark&-&6.75&9.98&7.34&10.86&7.32&10.86&6.86&10.22\\
	\multicolumn{1}{c|}{}&\checkmark&\checkmark&\checkmark&\textbf{6.62}&\textbf{9.74}&\textbf{6.99}&\textbf{10.23}&\textbf{6.88}&\textbf{10.20}&\textbf{6.78}&\textbf{10.03}\\

  \bottomrule
  \end{tabular}
\end{center}
\end{table}

\begin{table}[h]\renewcommand\tabcolsep{8pt}
\begin{center}
\newcommand{\tabincell}[2]{\begin{tabular}{@{}#1@{}}#2\end{tabular}}
  \caption{Number of Parameters with or without PoI-MetaBlock.}
  \label{tab:result4}
  \begin{tabular}{c|ccccc}
  \toprule
    \multicolumn{1}{c|}{\tabincell{c}{POI-\\MetaBlock}}&DCRNN&STGCN&ASTGCN&GAT-Seq2Seq&GMAN\\
    \hline
	-&0.07 M&0.03 M&0.08 M&0.04 M&0.06 M\\
	\checkmark&0.11 M&0.07 M&0.12 M&0.08 M&0.10 M\\

  \bottomrule
  \end{tabular}
\end{center}
\end{table}

\textcolor{black}{Our proposed POI-MetaBlock comprises three main components: dynamic attention, parameter generation, and attention refining. To evaluate the effectiveness of each component, we conducted experiments using some state-of-the-art models. Table \ref{tab:result3} shows the performance results, where "DA" indicates dynamic attention, "PG" represents parameter generation, and "AR" denotes attention refining.}

\textcolor{black}{Based on the results presented in Table \ref{tab:result3}, it is evident that each component of our proposed POI-MetaBlock (dynamic attention, parameter generation, and attention refining) contributes to the overall improvement in the performance of the baseline model. The improvement brought by dynamic attention confirms the high correlation between traffic patterns and time. The improvement brought by parameter generation indicates that traffic patterns in different regions are distinct, and generating parameters for each region enables the model to recognize these differences. Finally, attention refining achieves further improvements, validating that the functional attributes represented by POI data have a significant impact on traffic patterns that cannot be captured by the geographic topology alone.}

\textcolor{black}{We performed a detailed analysis of the number of parameters of the model. Table~\ref{tab:result4} shows the changes in the number of parameters of the model after adding PoI-MeteBolck. Parameters of PoI-MetaBolck are mainly composed of self-attention prediction and graph convolution refining.  The conventional methods use a single set of parameters to predict the traffic within all regions while PoI-MetaBlock generates independent parameters for each region. Compared with maintaining independent prediction parameters for each region, the parameter generation strategy allows the model to fit different traffic patterns in areas with different functions in a more lightweight manner.}

\subsection{Evaluation on Framework Settings}

\textcolor{black}{Our POI-MetaBlock has several hyperparameters, including the block's output dimension and the threshold of POI similarity for constructing the graph structure. To explore the impact of these settings on the model's performance, we conducted experiments using DCRNN under different module configurations.}

\begin{table}[t]\renewcommand\tabcolsep{10pt}
\begin{center}
\newcommand{\tabincell}[2]{\begin{tabular}{@{}#1@{}}#2\end{tabular}}  
  \caption{Effect of Different Output Dimensions.}
  \label{tab:outdim}
  \begin{tabular}{c|cc|cc}
   \toprule
    \multirow{2}*{\tabincell{c}{Output\\Dims}}& \multicolumn{2}{|c|}{Chengdu}&  \multicolumn{2}{|c}{Xi'an}\\
    \cline{2-5}
	\multicolumn{1}{c|}{}&MAE&RMSE&MAE&RMSE\\
	\hline
	\multicolumn{1}{c|}{8}&9.40&13.49&6.79&10.02\\
	\multicolumn{1}{c|}{16}&9.13&13.44&6.62&9.74\\
	\multicolumn{1}{c|}{32}&9.14&13.48&6.61&9.77\\
	\multicolumn{1}{c|}{64}&9.14&13.46&6.60&9.73\\
  \bottomrule
\end{tabular}
\end{center}
\end{table}

\begin{figure}[t]
  \centering
    \includegraphics[width=3.1in]{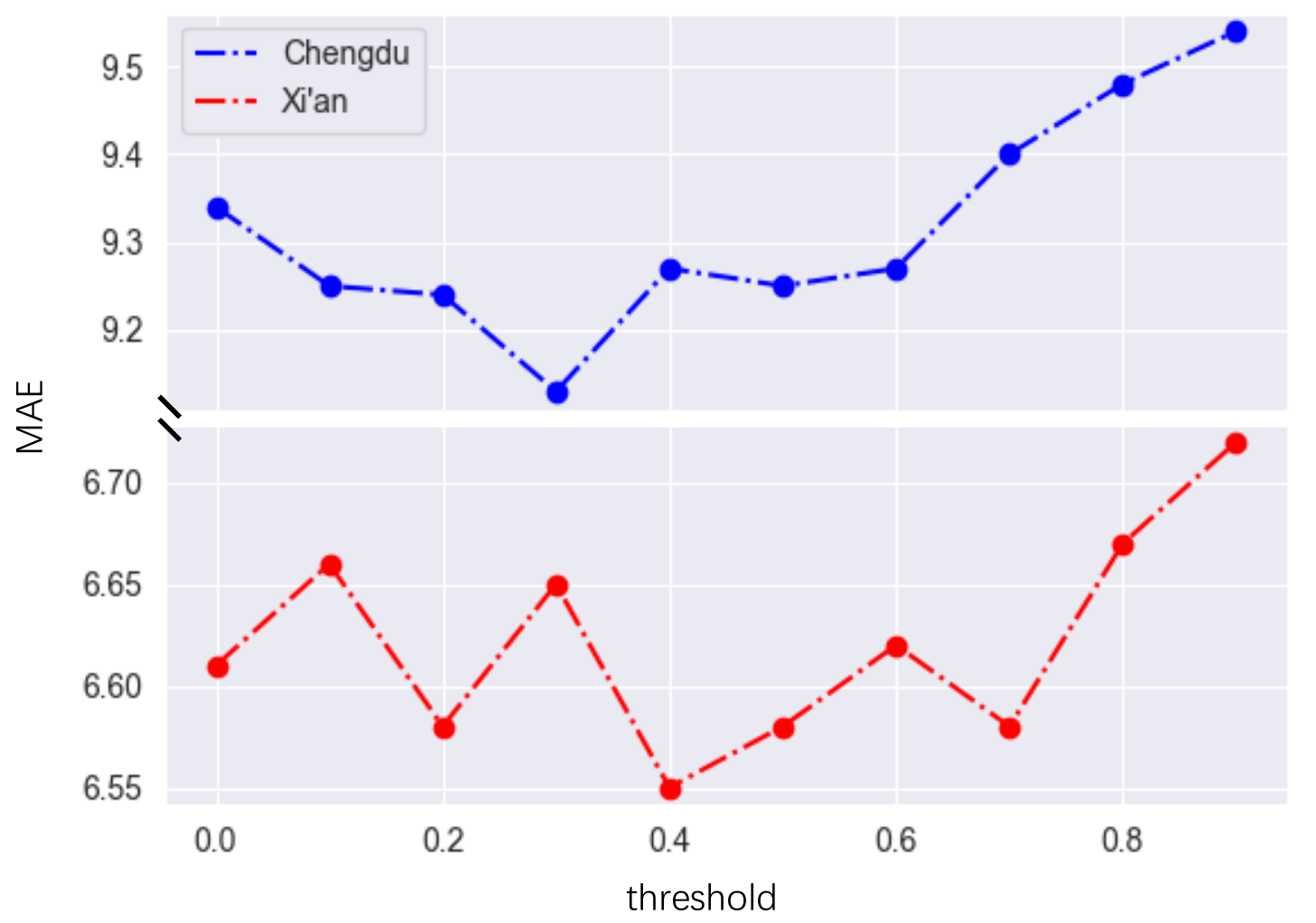}
  \caption{Effect of different thresholds.}
  \label{pic10}
\end{figure}

\textcolor{black}{Table \ref{tab:outdim} shows the effects of different output dimensions on both datasets, with the POI similarity threshold set to 0.4. We observed that an output dimension of 8 performs poorly, indicating that it is insufficient to represent complex traffic characteristics. The other output dimensions perform almost equally, and different output dimensions lead to different parameters, which affect the parameter generation process. The traditional method sets the output dimension as 32 and then uses the full connection layer to obtain prediction results. However, we found that an output dimension of 16 performs well enough and generates fewer parameters, resulting in faster model runtimes.}

\textcolor{black}{The effects of different POI similarity thresholds on the performance of the model are presented in Figure \ref{pic10}. In this experiment, we set the output dimension to 16. The threshold is used to construct the graph structure in the attention refining process. A smaller threshold generates more edges in the graph structure. From the results, we observe that the best thresholds for the two datasets are 0.3 and 0.4, respectively. Moreover, a smaller threshold does not necessarily result in better performance, indicating that connecting regions with different functional attributes can negatively affect the model's performance. Conversely, a high threshold of 0.8 and 0.9 results in lower performance due to insufficient connections between regions. Therefore, it is recommended to select a threshold in the range of 0.3 to 0.6 for better performance in the experiments.}

\section{Conclusion}

\textcolor{black}{Based on the premise that the functional attributes of urban areas have an impact on traffic flow, we propose to use POI data to represent these attributes and design a module called POI-MetaBlock to guide traffic flow prediction based on regional functionality. To verify the effectiveness of our proposed module, we collect trajectory and POI data and create two traffic flow datasets through irregular partitioning. We integrate our module into five traditional traffic prediction models and demonstrate through experiments on the two datasets that our module can significantly improve performance. Furthermore, we compare our module with a state-of-the-art model that utilizes meta information to assist prediction, and our experimental results show that our module outperforms the state-of-the-art methods.}

\bibliography{mybib}

\end{document}